\documentclass[acmtog]{acmart}
\acmSubmissionID{519}

\title{Catch \& Carry: Reusable Neural Controllers for Vision-Guided Whole-Body Tasks}

\author{Josh Merel}
\email{jsmerel@google.com}
\author{Saran Tunyasuvunakool}
\author{Arun Ahuja}
\author{Yuval Tassa}
\author{Leonard Hasenclever}
\author{Vu Pham}
\author{Tom Erez}
\author{Greg Wayne}
\author{Nicolas Heess}
\affiliation{%
  \institution{DeepMind}}

\renewcommand{\shortauthors}{Merel \textit{et al.}}

\usepackage{hyperref}
\usepackage{natbib}
\usepackage{graphicx}
\usepackage{xcolor}
\usepackage{amsmath,amsfonts,bm}

\DeclareMathOperator*{\argmin}{arg\,min}

\usepackage{booktabs} 
\usepackage{cuted}
\usepackage{capt-of}

\setcopyright{rightsretained}
\acmJournal{TOG}
\acmYear{2020}\acmVolume{39}\acmNumber{4}\acmArticle{39}\acmMonth{7} \acmDOI{10.1145/3386569.3392474}

\setcitestyle{nosort, square}

\usepackage{array}
\newcolumntype{S}{>{\centering\arraybackslash}m{2cm}}
\newcolumntype{L}{>{\centering\arraybackslash}m{2.75cm}}

\begin{document}



\begin{teaserfigure}
  \centering
  \includegraphics[width=\textwidth]{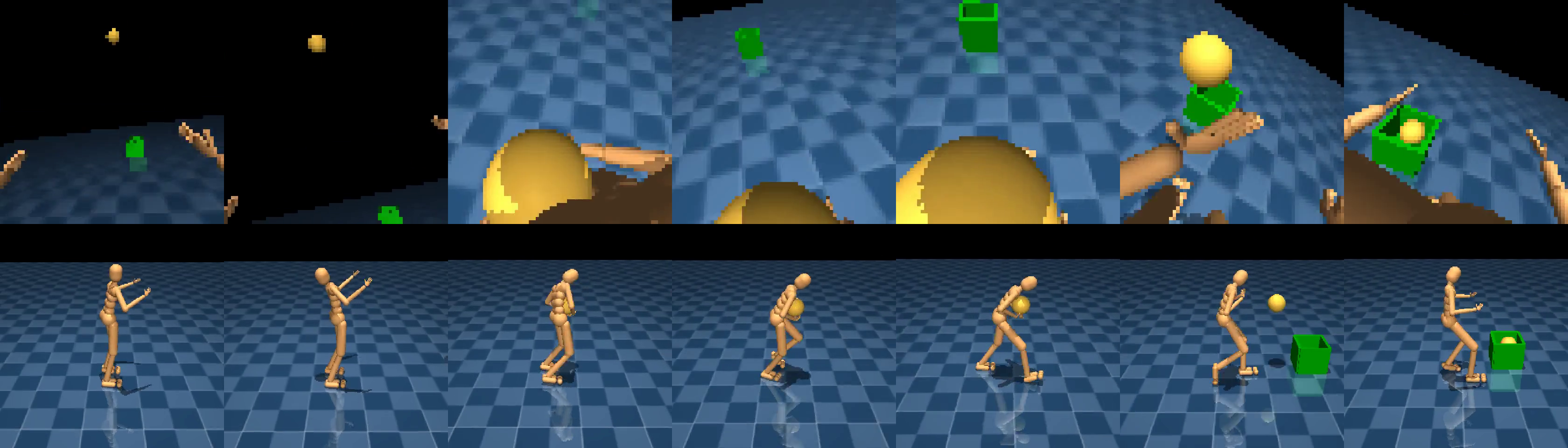}
  \caption{A catch-carry-toss sequence (bottom) from first-person visual inputs (top). Note how the character's gaze and posture track the ball.}
\end{teaserfigure}

\begin{abstract}

We address the longstanding challenge of producing flexible, realistic humanoid character controllers that can perform diverse whole-body tasks involving object interactions. This challenge is central to a variety of fields, from graphics and animation to robotics and motor neuroscience. Our physics-based environment uses realistic actuation and first-person perception -- including touch sensors and egocentric vision -- with a view to producing active-sensing behaviors (e.g. gaze direction), transferability to real robots, and comparisons to the biology. We develop an integrated neural-network based approach consisting of a motor primitive module, human demonstrations, and an instructed reinforcement learning regime with curricula and task variations. We demonstrate the utility of our approach for several tasks, including goal-conditioned box carrying and ball catching, and we characterize its behavioral robustness.  {The resulting controllers can be deployed in real-time on a standard PC.}\footnote{See overview video, {\color{blue} 
\href{https://youtu.be/2rQAW-8gQQk}
{\bf Video 1}}. Refer to Supplementary Section \ref{videos} for all video captions.}

\end{abstract}

\maketitle

\section{Introduction}

Endowing embodied agents with the motor intelligence that is required for natural and flexible goal-directed interaction with their physical environment is a longstanding challenge in artificial intelligence \citep{pfeifer2001understanding}. This is a problem of obvious practical relevance to a number of fields including robotics \citep{arkin1998behavior} and motor neuroscience \citep{merel2019hierarchical}. But it is also a topic of longstanding interest in the graphics and animation setting \citep[since e.g.][]{raibert1991animation, van1993sensor} -- the ability to control agents with physically simulated bodies and sensors that naturally behave in response to high-level instructions may reduce the effort in creating realistic animations of agent-environment interactions. 

Conventional approaches such as manual editing and kinematic blending of motion capture data require specification of character movements in their entirety, including how movements conform to the surrounding scene and task context. This can be challenging, especially when fine-grained motion is required, e.g. for object interaction. However, when controlling a goal-directed agent with physically simulated body, many aspects of its movements will emerge from the physical interaction; while other features, such as gaze direction, emerge from the interaction between the agent's goals and the constraints imposed by its body and sensors.

Recent developments in Deep Reinforcement Learning (Deep RL) have enabled great strides in learning from scratch for many game domains \citep{mnih2015human, silver2018general}; and while complex motor control problems involving physical bodies remain difficult even in simulation, there have been some successes discovering locomotion from scratch for reasonably sophisticated bodies \citep{heess2017emergence}. But generation of more complex behaviors, especially whole body movements that include object interactions, have remained largely out of reach.
These settings require the coordination of a complex, high-dimensional body to achieve a task goal, and satisfactory performance is often a complicated intersection of task success and additional constraints (e.g. naturalness, robustness, or energy-efficiency of the movements). Unlike problems with clear single-task performance objectives, these criteria can be hard to formalize as a reward function. Even where this is possible, the discovery of good solutions through RL can be difficult.  

Consequently, physics-based control algorithms often leverage prior knowledge, for instance, in the form of demonstrations or skills transferred from other tasks. These can help with the discovery of rewarding behavior \citep[e.g.][]{heess2016learning} as well as constrain the solutions that emerge \citep[e.g.][]{peng2018deepmimic,merel2018neural}.
These control settings force us to confront a fundamental trade-off:
while narrow, stereotyped skills, e.g. from demonstrations, 
can serve as useful initializations in settings where the controller only needs to reproduce one movement pattern, novel compositions of movements may be required for other settings, which, while related to demonstrations, are not completely consistent with them.
This is particularly pertinent to whole-body humanoid control that includes object interaction. While locomotion skills are only a function of the body's pose and its relation to the ground, manipulation skills are inherently tied to objects; yet we want manipulation skills to be general 
enough to apply not just a single scene with a particular object, but also to novel objects and object configurations.

Here we develop an integrated learning approach for humanoid whole-body manipulation and locomotion in simulation, that allows us to strike a satisfactory balance between task-specificity and motor generality for object interaction behaviors.  It consists of the following components: (1) a general purpose \textit{low-level motor skill module} that is derived from motion capture demonstrations, yet is scene agnostic and can therefore be deployed in many scenarios; (2) a hierarchical control scheme, consisting of a high-level \textit{task policy} that operates from egocentric vision, possesses memory, and interfaces with the the motor module; (3) a training procedure involving a broad distribution of \textit{task variations} to achieve generalization to a number of different environmental conditions; and lastly, (4) training using a \textit{phased task}, in which the task policy is trained to solve task stages using simple rewards, which, together with the use of demonstrations, greatly facilitates exploration and allows us to learn complex multi-step tasks while minimizing the need for complicated shaping rewards.  
{This approach builds on the \emph{neural probabilistic motor primitives} (NPMP) approach of \citealt{merel2018neural}. While that work demonstrated reusable skills for locomotion behaviors without objects, the present work adapts the training procedure to produce a similarly structured motor skill module from demonstrations that can successfully be reused for varied object interactions. In particular, to support invariance and transfer of motor skills across type and quantity of objects, the low-level controller is not provided direct access to object state, even while trained from demonstrations containing objects.}

We apply our approach to two challenging tasks, both involving a humanoid interacting bimanually with large objects such as boxes and medicine balls. The two tasks are an instructed box manipulation task in which the simulated character needs to follow user-specified instructions and move boxes between shelves (a simplified ``warehouse'' setting) and a ball catching and tossing task (``toss''). Both tasks are solved either from task features or egocentric vision by the same motor module (albeit different task policies) demonstrating the possibility of general and reusable motor skills that can be deployed in  diverse settings. 
The goal directed training leads to the robust deployment of locomotion and manipulation skills as well as active control of the head for gaze direction. 
The results demonstrate the flexibility and generality of the approach, which achieves significant generalization beyond the raw demonstrations that the system was bootstrapped from, and constitute another step towards general learning schemes for sophisticated whole-body control in simulated physical environments.

\section{Related work}

Research aiming to achieve coordinated movement of virtual humanoids occurs in artificial intelligence and physics-based character animation, and advances in these fields relate to developments in robotics and biological motor control.

\paragraph{Virtual humanoid control}
Realistic animation of humanoid characters is a major, enduring aim in the graphics community. {For generalized locomotion, various kinematic approaches sequentially set the pose of the body according to motion capture snippets \citep{lee2002interactive, arikan2002interactive, kovar2008motion}, with behavioral sequencing resulting from approaches ranging from high-level planning \citep{levine2011space} to timestep-level control by neural network sequence models \cite{holden2017phase}.}  
A parallel line of research, that we focus on here, employs physics-based simulation to control a virtual body and produce realistic movement \citep[e.g.][]{van1993sensor, Faloutsos2001}, often also using motion capture as a reference for the controller  \citep[e.g.][]{yin2007simbicon, liu2010sampling, peng2018deepmimic}.  For both kinematic and physics-based approaches, the aim is essentially to re-sequence or schedule the available movements to produce appropriate combinations; this is adequate for settings that require limited movement diversity.  In the context of dynamic locomotion, impressive results have been obtained by sequencing a small set of movements \citep{liu2012terrain}.  
Complementary to the use of motion capture, Deep RL approaches enable learning from scratch \citep{peng2016terrain, heess2017emergence}, with behavior emerging from the constraints of the physics and task. Task-driven learning can also occur in combination with motion capture which constrains behavior to be more humanlike \citep{peng2017deeploco, peng2018deepmimic, merel2017learning}.
{Recent approaches aim to more tightly couple kinematic and physics-based approaches by using kinematic approaches to specify  targets for physics-based controllers \citep{chentanez2018physics, park2019learning, bergamin2019drecon}.}

While locomotion is increasingly tractable in graphics settings using either kinematic or physics-based approaches, coordinated locomotion and manipulation remains more challenging.
Recent kinematic-based approaches show impressive interactions between a humanoid and the environment \citep{starke2019neural}, but this requires modeling the motion as an explicit function of a high-dimensional volumetric scene context. For highly dynamic movements the precise coupling between motion and scene elements can be difficult to model accurately \citep{liu2018learning}. 
{For many physics-based graphics approaches involving object interactions, a shortcut is often taken by forming fixed attachments between the hands of the body and the manipulated object in order to simplify the problem \citep{coros2010generalized, mordatch2012discovery, liu2012terrain, peng2019mcp}. Nevertheless, in addition to robotics-oriented efforts \citep[e.g.][]{sentis2005synthesis, otani2017adaptive}, there are instances of physically simulated object interactions, such as a humanoid sitting in a chair \citep{chao2019learning}.  Another particularly impressive exception involves dribbling a basketball simulated with physically plausible object interactions \citep{liu2018learning}; however, this work, while impressive, produces a dribbling controller that follows a relatively narrow trajectory.  
Narrow behaviors may be appropriate for certain animation settings, or even medical applications in which musculo-tendon models can help characterize the impact of surgical interventions \citep{holzbaur2005model, lee2019scalable}.
But generality of behavior may be important for settings that prioritize autonomy, such as autonomous virtual characters or robotics settings.
}

{Finally, we note that a necessary component of an integrated embodied system is control of head and gaze.  The animation and graphics literature has put considerable effort into making the eye movements of virtual characters appear natural \citep{ruhland2015review, pejsa2016authoring}, including by recent data-driven approaches \citep{klein2019data}. In the context of the present work, the active gaze control serves a functional role in a visuomotor task, and body movements are informed by inferred state \citep[see also][]{terzopoulos1995animat, sprague2007modeling}.  In graphics research, this kind of gaze control has been considered, for example, to enable catching behaviors requiring upper body movement \citep{yeo2012eyecatch, nakada2018deep, eom2019model} as well as visually guided locomotion \citep{eom2019model}, generally using specially engineered gaze-control systems.
In our specific setting, we do not model eyes, but the agent can learn to control its gaze via head movements in an emergent fashion in order to support task performance.}

\paragraph{Generalizing from demonstrations}
Motion capture for virtual character control in graphics is one particular instance of the more general problem of leveraging demonstrations for control.
Demonstrations can be readily obtained for many simple real or simulated robotics systems, for instance through \textit{teleoperation} or via a human operator physically guiding the pose of the robot.
The classical approach for learning from demonstrations amounts to using the demonstration to initialize the policy, and learning how to deviate from the demonstrate to solve the task at hand \citep{smart2002effective, schaal2003computational}.  It has long been recognized that given a small number of demonstrations, it is not sufficient to try to directly mimic the demonstration as there will be some discrepancies when recapitulating the movement which will compound and lead to failure \citep{atkeson1997robot, schaal1997learning}.  A fairly direct approach involves fitting the demonstrations to a parametric form and using RL to modulate the parameters of the fitted model \citep{guenter2007reinforcement, peters2008reinforcement, kober2009policy, pastor2011skill}. Slightly more indirect approaches consist of using the demonstrations to learn local models from which a policy can be derived \citep{coates2008learning} or using the demonstrations to infer the objective for the policy through inverse optimal control \citep{ng2000algorithms, ho2016generative, englert2018learning}.  In Deep RL settings involving a replay buffer, and when the demonstrations include actions and reward on the task being solved, it is possible to fill the replay buffer with teleoperation demonstrations \citep{vevcerik2017leveraging}.  Finally, as noted previously, there are approaches in which both matching the demonstrations and solving the task serve as rewards during training \citep{kumar2016learning, peng2018deepmimic, merel2017learning, zhu2018reinforcement}.  Note that insofar as the main role of demonstrations is to serve as a form of prior knowledge, a similarly motivated approach is to design controllers that incorporate domain knowledge for some tasks and to then use learning to refine the behavior around this initial, engineered policy -- this scheme has been applied to efforts in robotics for tossing objects \citep{zeng2019tossingbot} and catching objects \citep{kim2014catching}.

\begin{figure*}[ht]
\centering
\includegraphics[width=1.\linewidth]{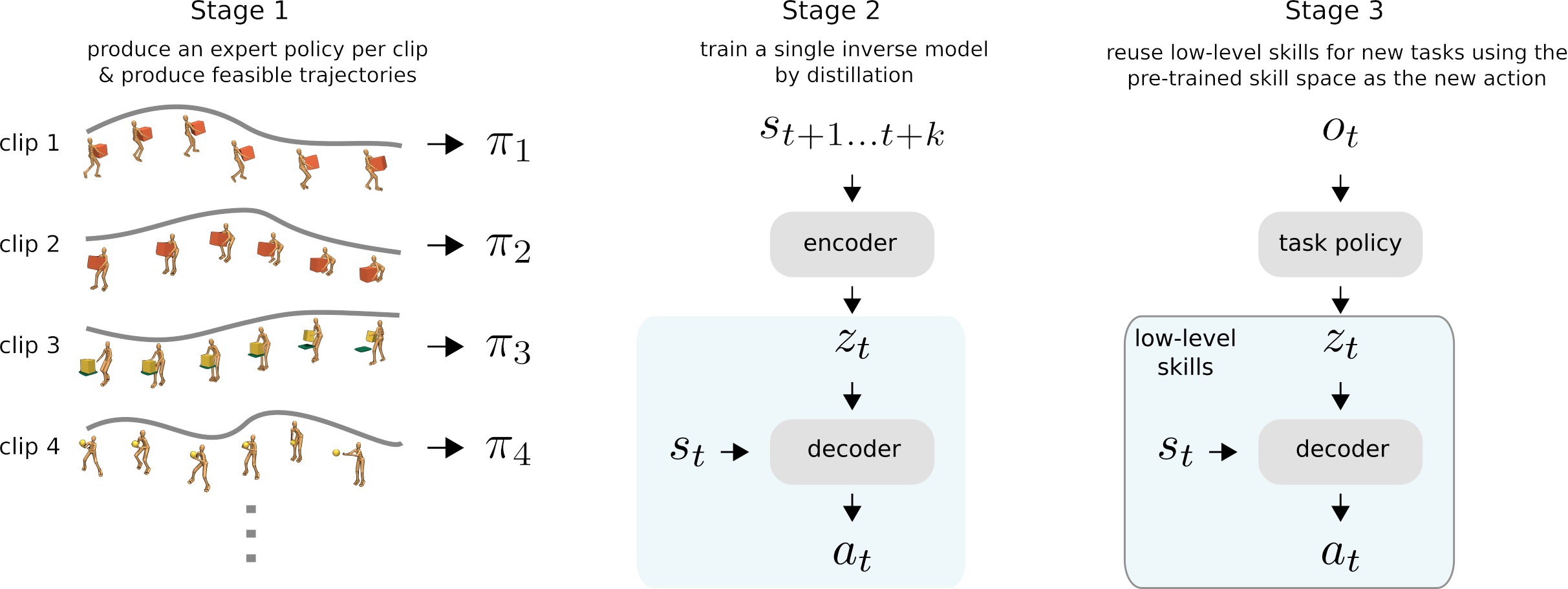}
\caption{{\bf Overview of producing and reusing skills.}  Stage 1 involves training a large set of separate, single-behavior ``expert'' policies.  For each motion capture clip trajectory (depicted by a curve), we produce a policy that tracks that trajectory. Stage 2 involves distilling the experts produced in stage 1 into a single inverse model architecture. The inverse model receives the state for a few future steps (from $t+1$ to $t+k$ where $k=5$ in our setting), an encoder embeds this into a latent intention ($z_t$), and the decoder produces the action that will achieve the transition from $s_t$ to $s_{t+1}$.  Stage 3 involves training only the task policy to reuse the frozen low-level skill module, using the learned embedding space to communicate what to do.}
\label{fig:overview}
\end{figure*}

The commonality across this broad class of existing approaches for learning from demonstrations, both in virtual and robotic settings, is that they are well suited primarily when there is a single variety of movement that needs to be reproduced and where the demonstrations are well aligned with the behavior required to solve the task.  While these approaches have been successful in various cases, they have not yet been demonstrated for more complex tasks that require composition and arbitrary re-sequencing of motor skills.  
Ideally we wish for a skill space to serve both as a generic ``initialization'' of the policy as well as a set of constraints on the behavior; yet we also want the skill space to be multipotent, in the sense that it can be leveraged for multiple distinct classes of target tasks, rather than serve only for a narrow range of movements.
While some work has aimed to build motor skill modules from unstructured demonstrations \citep{jenkins2003automated, niekum2012learning}, limited work to date has aimed to learn flexible skill modules of the sort suitable for Deep RL \citep{merel2018neural, peng2019mcp}.  It remains open how best to generalize beyond individual trajectories and strike a balance between realism of the movements and the degree to which new movements can be synthesized from finite demonstrations.

\section{Approach}
In this work, we develop an approach for skill transfer and learning from demonstrations in the setting of visually-guided humanoid control with object interactions.  By ``skill transfer'', we refer to a setting which involves establishing basic motor competency on a source distribution of demonstrations or tasks and applying this prior knowledge to improve learning on new tasks.  
{Our approach employs an episodic reinforcement learning (RL) paradigm, in which tasks are defined for a physical scene by specifying initial conditions, termination conditions, and a task-specific per timestep reward ($r_t$).  A learning agent maintains a policy $\pi$ that produces actions ($a_t$) in response to partial observations that it receives of the underlying physical state ($s_t$) of the environment. The agent aims to learn parameters of the policy so that it maximizes the expected sum of discounted future rewards during rollouts from the policy, $\mathbb{E}_{\pi}[\sum_{t=0}^{\infty} \gamma^t r_t]$.
We propose and evaluate an approach that leverages unlabelled demonstrations (without rewards) and creates a \emph{motor module} or low-level controller, which can then be used for multiple object-interaction tasks. This approach can also be seen as incorporating additional, previously learned structure into the policy to facilitate subsequent learning. We emphasize that while training a policy from scratch on motor problems is difficult insofar as RL-based policy optimization only obtains local optima, approaches that constrain the search space can result in more tractable optimization and yield better solutions.}  

The general workflow for producing and reusing the skill module is depicted in figure \ref{fig:overview}.  
It consists of three stages.  Firstly, expert neural-network policies are generated which are capable of robustly tracking individual motion capture clips in the presence of noise.  The second stage consists of distilling these policies into a single conditional policy, or inverse model, which maps the state at the current timestep ($s_t$) and the desired state at timesteps in the near future ($s_{t+1...t+k}$) to the first action ($a_t$) of a sequence of actions that would result in that desired future.  
{Providing a horizon into the future (i.e., $k>1$, as opposed to only providing $s_{t+1}$) is potentially useful in disambiguating the instruction for movements that require multiple timesteps of preparation to perform.  As explained in more detail in Section \ref{sec:approach:npmp}, this inverse model is separated into an encoder and decoder, which communicate via a multi-dimensional, continuous random variable that reflects short term motor intention. The decoder can also be interpreted as a conditional policy that is trained via a form of behavioral cloning. This training procedure generally follows the approach of \citep{merel2018neural}, and so we refer to this architecture as ``Neural Probabilistic Motor Primitives'' (NPMP).  Finally, the third stage amounts to reusing the NPMP decoder as a low-level controller in the context of new tasks, by treating the learned motor intention space as an action space for a new controller. Here, a high-level task policy receives observations that are appropriate for the target task, either vision or states of relevant objects relative to the body, and this policy is trained by model-free RL.  The task policy outputs ``actions'' corresponding to latent variables that serve as ``commands'' to the now fixed low-level module. The low-level module thus transforms the initial noise distribution of an untrained task policy into ``colored''-noise that reflects the coordinated movement statistics of the motion capture data.
By acting through the low-level controller, movement exploration as well as resulting solutions are constrained to the manifold of human-like behavior that can be produced by the motor module.}

The particular challenges of the manipulation tasks considered in this work mean that several additional elements of the training process are critical. Manipulation requires directed interaction with objects in environment, and these are difficult to discover even when exploration is restricted to the space of movements expressed by the skill module. We address this by employing a suitable distribution of initial configurations of the body and objects, along with variations for object masses and sizes.
Taken together, these initializations and variations facilitate learning through exposure to scene configurations that vary in difficulty and distance from reward, offering an organic curriculum.
Finally, while we find that the skill module is somewhat robust to variations in which expert demonstrations are included; there is a trade-off between specificity and generality of skills.  We elaborate on these elements of our proposed training process when presenting the tasks, and we demonstrate their relevance through ablations (see Results, Section \ref{sec:ablations}).

\begin{figure}[b]
\centering
\includegraphics[width=.95\linewidth]{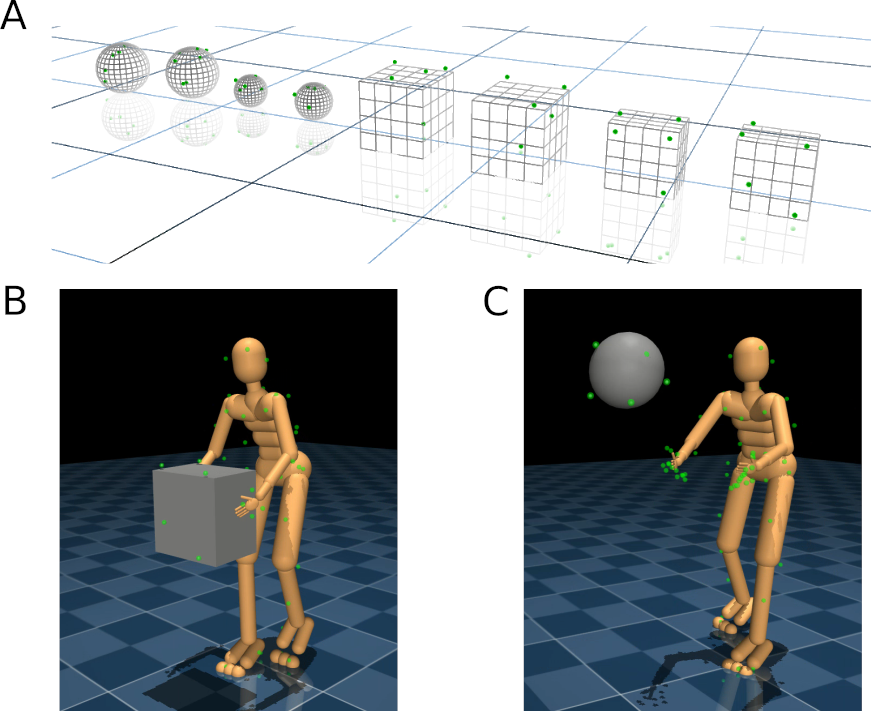}
\caption{{\bf Motion capture.} (A) Virtual analogs of the objects that are tracked with motion capture.  (B) \& (C) Frames of motion capture for box interaction and ball tossing, with prop and humanoid body set to those poses in the physics simulator. The green dots correspond to markers.}
\label{fig:demos}
\end{figure}

All simulations are performed using the MuJoCo physics simulator \citep{todorov2012mujoco}.  The humanoid body has 56 actuated degrees of freedom, 
and was adapted from the standard version of this body that is available as part of the DeepMind Control codebase \citep{tassa2018deepmind}.  
{Similarly to previous work using this body, we use position-control actuators that are limited to produce reasonable maximum torques. The body also includes force sensors in the shoulder as well as multiple binary touch/contact sensors on each hand. For this work, all body lengths are scaled from the available body according to the measured dimensions of the motion capture subject who performed reference movements.}  

\subsection{Demonstrations for skills}

We collected motion capture data of a person performing bimanual, whole-body box manipulation movements, ball tossing, and various locomotor behaviors with and without objects in-hand.  When objects were involved, we also collected motion capture for the objects.  To go from point-cloud, raw motion capture data to body-specific movements, we implemented simultaneous tracking and calibration \citep{wu2013stac}, which solves a joint optimization problem over body pose and marker position.  See Supplementary Section \ref{stac_details} for implementation details.  Figure \ref{fig:demos} shows a visualization of the virtualized ``props'' and humanoid set to poses from the motion capture.  As noted above, it was important that we measured and re-sized the lengths of body segments of the virtual character to correspond to the person whose motion data we collected, as this ensured that the positions of the hands relative to tracked objects are similar in the virtual environments relative to the real setting. This precision in body dimensions also made the inference of body poses from point-clouds more robust.  Nevertheless, the proportions of the virtual humanoid still only approximately correspond to the human actor and the dynamic properties differ substantially. 

The dataset collected for this work consists of a single subject interacting with 8 objects (or ``props'').  The objects are two ``large'' balls, two ``small'' balls, two ``large'' boxes, and two ``small'' boxes. Small objects weighed 3kg and large objects 10kg.  We considered interactions at 3 heights, ``floor-height'', ``torso-height'', and ``head-height''.  For each object, at each height, we collected two repeats of behavior consisting of the actor approaching a pedestal on which an object is resting, picking it up, walking around with the object in hand, returning to the pedestal, placing the object back on the pedestal, and then backing away from the pedestal.  In total this amounts to 48 clips (8 objects $\times$ 2 repeats $\times$ 3 heights), each of which is generally no less than 10 seconds and no longer than just over 20 seconds. 
Other less structured behavior was captured, including walking around with no object (``walking'') as well as tossing a ball with a second person (``ball-tossing''; one person and the ball were tracked).  In total, we use a little less than 20 min of data ($\sim$1130 sec).  For representative examples, see videos of motion capture  playback: box interaction 
{\color{blue} 
\href{https://youtu.be/myRC8kAcftY}
{\bf Video 2}} and ball tossing 
{\color{blue} 
\href{https://youtu.be/W_V2lb82qjo}
{\bf Video 3}}.

\subsection{Single-clip tracking for object manipulation}

\begin{figure}[]
\centering
\includegraphics[width=\linewidth]{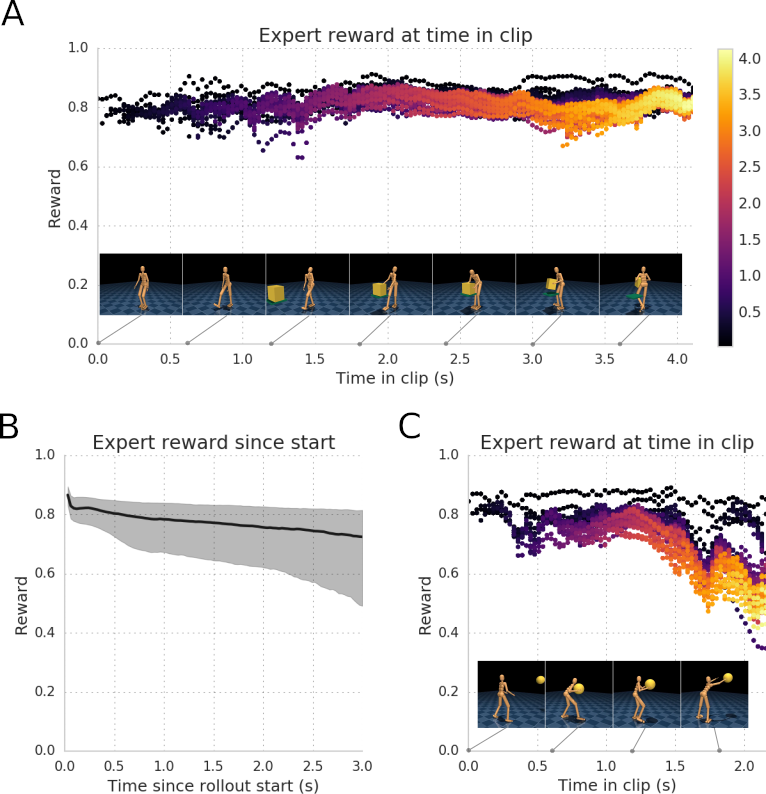}
\caption{{\bf Tracking with expert policies.} (A) Tracking performance and corresponding filmstrip is shown for a representative warehouse expert clip. We initialize the expert at timepoints throughout the clip and the policy controls the behavior to end of the clip. Time within a rollout since the initialization is depicted via intensity. The policy is robust in that it controls the body (interacting with the box) to remain on track. (B) Representative summary of the expert tracking performance for rollouts as a function of time since initialization for all medium height experts -- rollouts lasting up to 3 seconds show only limited accumulated tracking error, indicating experts are well-tracked.  (C) Performance and filmstrip for a ball-tossing expert indicating good tracking performance until the ball is released from the hands at which point the performance deteriorates due largely to the loss of control over the ball. Despite the inability to control the ball to match the reference ball trajectory perfectly, visually, the expert looks reasonable through the release of the ball.}
\label{fig:expert_analysis}
\end{figure}

To produce expert policies, we use a tracking objective and train time-indexed policies to reproduce the movements observed via motion capture \citep{peng2018deepmimic}, here including the position of the object. Similarly to \cite{merel2018hierarchical}, we provide the agent a normalized tracking reward, $r_t \in (0,1]$, that reflects how well the body and object in the virtual environment match the reference:
\begin{equation}
\label{reward_eqn}
    r_t = \exp(-\beta E_{\rm total}/w_{\rm total})
\end{equation}
where $w_{\rm total}$ is the sum of the per energy-term weights and $\beta$ is a sharpness parameter ($\beta=10$ throughout).  The energy term is a sum of tracking terms, each of which corresponds to a distance between the pose of the physically simulated body relative to a reference trajectory derived from motion capture: 
\begin{align}
\label{energy_eqn}
    E_{\rm total} = &w_{\rm qpos}E_{\rm qpos} + w_{\rm qvel}E_{\rm qvel} + w_{\rm ori}E_{\rm ori} + \nonumber \\ &w_{\rm app}E_{\rm app} + w_{\rm vel}E_{\rm vel} + w_{\rm gyro}E_{\rm gyro} + w_{\rm obj}E_{\rm obj}
\end{align}
with terms for tracking the reference joint angles ($E_{\rm qpos}$), joint velocities ($E_{\rm qvel}$), root quaternion ($E_{\rm ori}$), body-frame vectors from the root to appendages (hands, feet, head; $E_{\rm app}$), translational velocity ($E_{\rm vel}$), root rotational velocities ($E_{\rm gyro}$) and object position ($E_{\rm obj}$). See Supplementary Section \ref{tracking_details} for more specific details.  Note that to encourage robustness of the controller, we train in the presence of moderate action noise -- noise is sampled from a Gaussian independently per actuator with $\sigma=.1$, for actions $\in [-1,1]$.  

\begin{figure*}[t!]
\centering
\includegraphics[width=.9\linewidth]{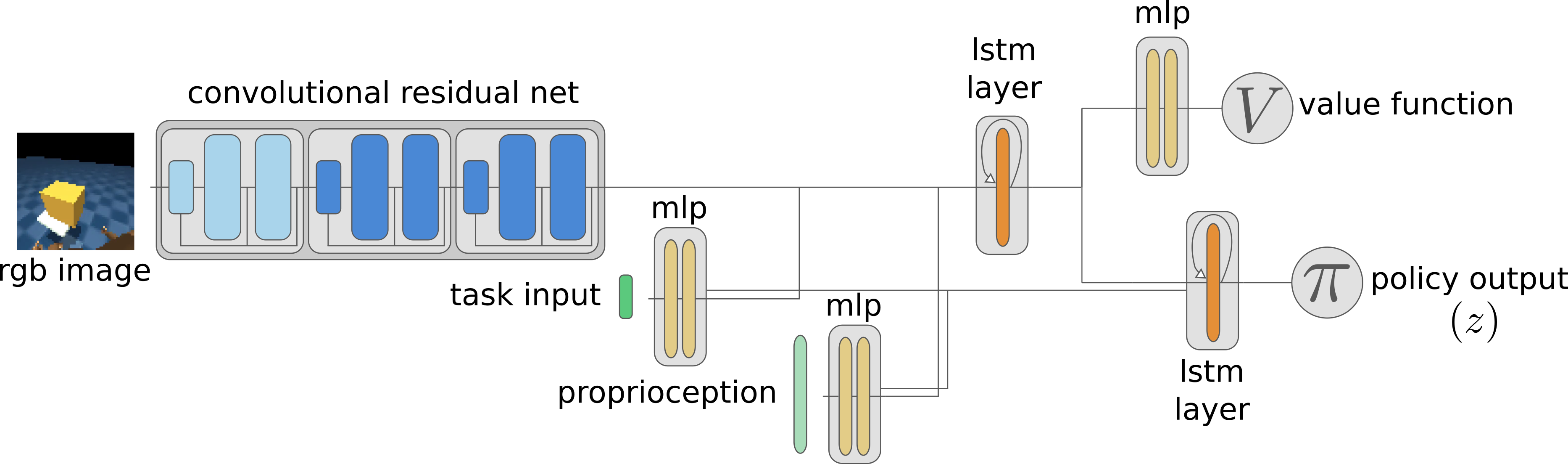}
\caption{{\bf High-level task policy architecture.} When training a high-level task policy to reuse the low-level controller, three streams of input are available potentially.  For whichever of the egocentric image input, task instruction input, and proprioception streams are available, each is passed through a preprocessor network. A value function and policy branch from a shared LSTM, and the policy also receives skip connections for the task and proprioception input streams.  The policy output here refers to high-level actions that serve as inputs to the low-level controller.}
\label{fig:task_policy_architecture}
\end{figure*} 

{Using the objective described above, we produce expert policies for all reference motions (at similar sample complexity to previous work).  Each expert is trained to track motion capture clips that are roughly 3-5s snippets of the raw reference clips, yielding a few hundred varied experts that tile the behaviors discussed in the previous section.
} To asses performance of the expert tracking controllers, we collect rollouts starting from different points along the trajectory and see how well these align with the motion capture reference.  In general we find that tracking performance is good for ``warehouse'' behavior experts (Figure \ref{fig:expert_analysis}A) with only a small falloff as a function of duration of rollout (Figure \ref{fig:expert_analysis}B).  For ``toss'' behavior experts, performance sometimes shows a sharp fall-off after tossing the ball (Figure \ref{fig:expert_analysis}C).  However, this performance decline is primarily due to the object tracking term when the ball is no longer directly controlled and does not reflect failure of body tracking, as visually discernible from the filmstrip in Figure \ref{fig:expert_analysis}C.  As a data augmentation, we also produced ``mime'' experts for which the expert was trained to track the human reference movements involving object interactions, but for which the object was not present in the virtual environment.
Inclusion of these mime experts helped balance the data used for distillation, which otherwise overrepresented object carrying movements, resulting in controllers that are overly predisposed to bring the hands of the humanoid together.

\subsection{Training the motor module for locomotion and manipulation}
\label{sec:approach:npmp}

The single-behavior expert policies track individual motion capture trajectories but do not directly generalize to new tasks or even configurations of the environment. 
To enable reusability of the skills, we therefore follow \citep{merel2018neural} and distill expert behaviors into a single module with suitable architecture (the ``Neural Probabilistic Motor Primitives'', or NPMP). 
Unlike \citep{merel2018neural} we are interested in manipulation skills which depend strongly on the environment, not just the body controlled by the agent.
As a critical design choice that ensures usability of the motor module across various environments, we employ the following factorization:  
during training, we give the encoder access to the state both of the humanoid as well as the object used in the expert trajectory; however, the decoder only directly receives egocentric humanoid proprioceptive information.
By construction, the decoder will therefore be reusable as a policy that only requires egocentric observations of the humanoid body, which are inherently consistent across environments. When reusing the skill module, any awareness of the objects in the scene must be passed to the low-level controller via the latent variable produced by the task policy.  

The training procedure for the motor module follows the approach presented in \cite{merel2018neural}. We train the model in a supervised fashion to model state-action sequences (trajectories)  generated by executing the various single-skill experts policies, while adding independent Gaussian noise to the actions.
Specifically, we maximize the Evidence Lower Bound (ELBO):
\begin{align}
    \label{eqn:ELBO}
    \mathbb{E}_q
    \Bigl[
    \sum_{t=1}^{T}\log \pi(a_t|s_t, z_t) &+ \beta \big( \log p_{z}(z_t|z_{t-1}) \nonumber \\
    &- \log q(z_t|z_{t-1}, s_{t+1...t+k})\big)
    \Bigr]
    ,
\end{align}
where $\pi(a_t|s_t, z_t)$ corresponds to the decoder, which is a policy conditioned on the latent variable $z_t$ and the current state $s_t$. The distribution 
$q(z_t|z_{t-1}, s_{t+1...t+k})$ corresponds to the encoder and produces latent embeddings based on short snippets into the future.
The latent variable $z_t$ carries the semantics of motor intention, inherited from training on the behavior distribution of the experts, and this space is therefore called the \emph{skill embedding} space. This latent $z$ will now serve as a task-relevant instruction to the decoder-policy.
$\beta$ controls the weight of the autoregressive prior $p_{z}(z_t|z_{t-1})$, which regularizes the skill embedding space by encouraging $z$'s temporal continuity.  
For further analysis of the skills well-reflected in this space, see Supplementary Section \ref{one_shot}.

\subsection{Training task policies that reuse low-level skills}
To train the task policy, which reuses the low-level skill module, we use a model-free distributed RL setup with a single learner and many actors (here 1000).
This training paradigm as well as our use of off-policy correction via V-trace to train the the value-function are consistent with IMPALA \citep{espeholt2018impala}. 
Policy updates are perfomed using a version of V-MPO \citep{song2019v}, which is in the MPO family of approaches \citep{abdolmaleki2018maximum}; however, while V-MPO proposes the use of on-policy updates, we perform updates from a replay buffer, which we found stabilized learning in our setting.
{Given these choices, the relevant learning parameters are the learning rate (1e-4), the MPO $\epsilon$ that controls the KL-divergence constraint on the policy (we swept .5 and 1.), and discount factor ($\gamma=0.99$).  In addition, the policy and value functions are updated with minibatches of size 128 and trajectory length of 50, from a replay buffer.}

The architecture we used for the task policies is close to the simplest architecture that is suited for the setting (see Figure \ref{fig:task_policy_architecture} for a schematic). We separately encode the qualitatively different streams of inputs with separate preprocessing networks -- small 1-2 hidden-layer MLPs for the task and proprioceptive features and a generic ResNet for image inputs \citep{he2016deep}.  The outputs of these three input channels are concatenated and used as inputs to an LSTM \citep{hochreiter1997long}. The shared LSTM branches into the value function output as well as a second LSTM which is used for the policy.  By having the first shared LSTM, learned representations that are useful both for the value function and policy can be shared.
We did not extensively tune the network structure.

We want the task policy to produce actions consistent with the values seen by the low-level skill module during its supervised training (the embedding space is regularized to have values close to zero).  To prevent actions from the task policy from being too ``out-of-distribution'' for the pretrained low-level controller, we restrict the high-level actions to a limited range of values, in the range $(-2, 2)$.

\section{Results}

\subsection{Core tasks}

In this work, we defined two challenging object interaction tasks, and we show that the low-level skill module can be used to solve either of these, when a high-level, task-specific policy is trained to reuse the skills on each task.  Our two core tasks are a proto-warehouse task (``warehouse'') and a ball tossing task (``toss'').  The warehouse task involves going to a box that is on a pedestal, picking up the box, bringing it to another pedestal, putting the box down, and repeating.  To make the task unambiguous (e.g. whether the current goal is to pick up or put down), we provide the agent with a task ``phase'' or ``instruction'' that indicates which of these four phases of the task the agent is presently in.  This phase also provides a natural way of providing sub-goals, insofar as sparse rewards are provided after each phase of the task has been completed.  In addition, we provide the agent with the position (relative to itself) of the target pedestal (to which it must go in order to pick up a box, or to put down a box).  
This active pedestal is highlighted in the videos, and this visual cue is also available to vision-based agents.  See Supplementary Section \ref{warehouse_supp} for further details about the task specification. 

Our second task consists of catching a ball and then tossing it into a bucket.  In this task, the ball is always initially thrown towards the humanoid.  The task is terminated with a negative reward if the ball touches the ground, which incentivizes the agent to learn to catch the ball and avoid dropping it.  A small shaping reward encourages the agent to bring the ball towards the bucket, and a sparse positive reward is provided if the ball is deposited into the bucket.  See Supplementary Section \ref{toss_supp} for details.

\begin{figure}[]
\centering
\includegraphics[width=1.\linewidth]{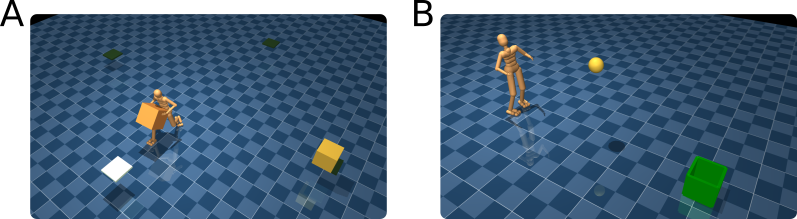}
\caption{{\bf Tasks.} (A) The ``warehouse'' task involving instructed movements of indicated boxes from one indicated pedestal to another.  (B) The ``toss'' task involving catching a ball thrown towards the humanoid and tossing it into a bucket on the ground.  Tasks can be performed either from state features or from egocentric vision.}
\label{fig:tasks}
\end{figure} 

{Both tasks are generated procedurally, with several task parameters sampled from a distribution on a per-episode basis.  For the warehouse the pedestal heights, box dimensions, and box masses are each sampled from a distribution.  For the tossing task, the ball size, mass, the trajectory of the ball thrown towards the humanoid, and the position of the bucket are each sampled from a distribution.  In both tasks, mass variations are visualized by object color (darker is heavier).  In the warehouse task, we also initialize episodes in the various phases of the task and sample initial poses of the body from the motion capture data.   Again, see Supplementary Sections \ref{warehouse_supp} and \ref{toss_supp} for more details. These task variations and initializations are important for successful training of the task policies as we will show in Section \ref{sec:ablations}. To observe the environment, agents are provided either visual information (an egocentric camera mounted on the head of the humanoid) or state features which consist of the position of the prop relative to the humanoid as well as the orientation of the prop, and we compare performance using these different observations.}  

\subsection{Performance on tasks}

\begin{figure}[b]
\centering
\includegraphics[width=1.\linewidth]{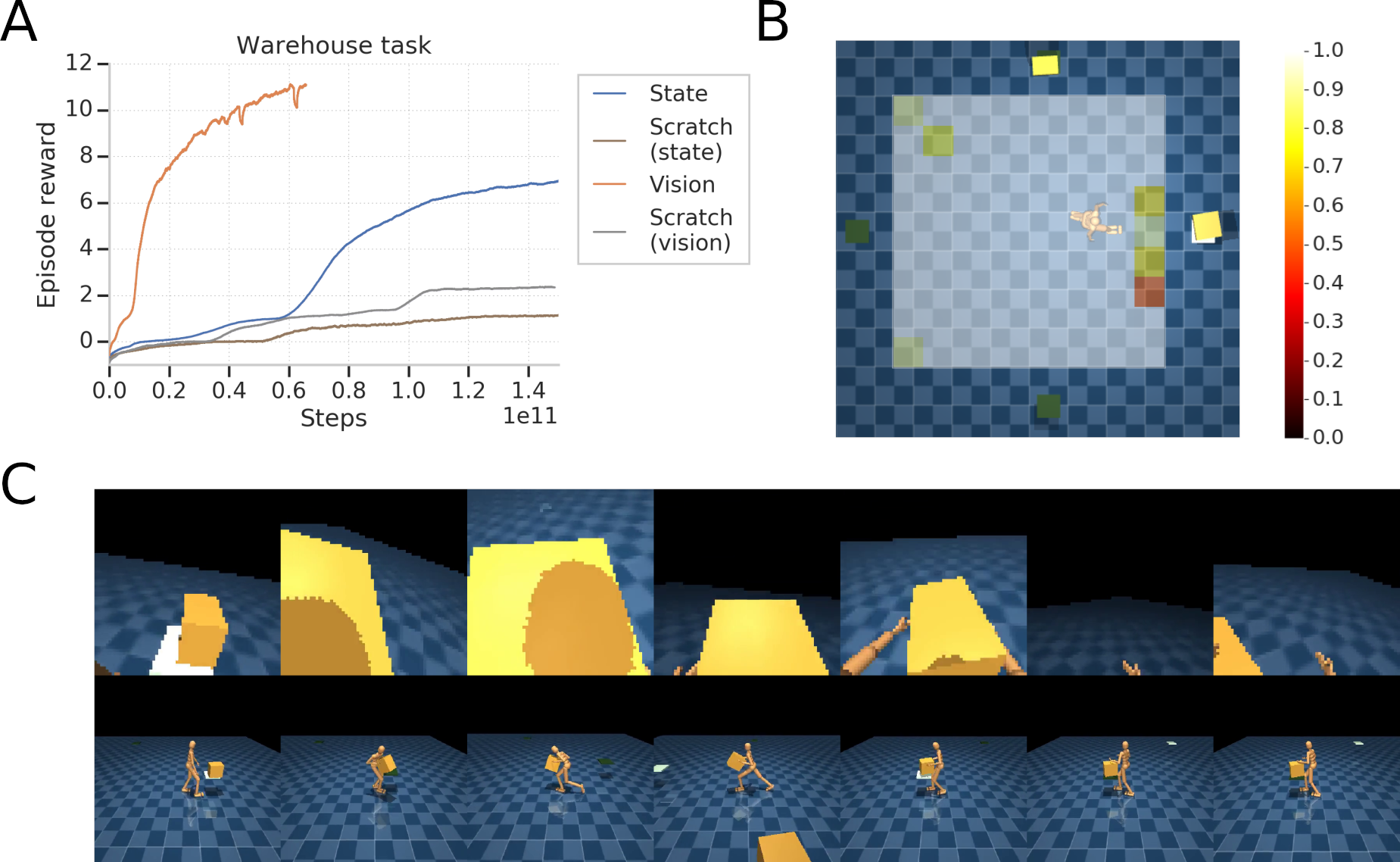}
\caption{{\bf Performance for the ``warehouse'' task:} (A) Representative learning curves (best of 3 seeds) comparing vision-based and state-based performance on the warehouse task, as a function of learner update steps.
(B) For the trained vision-based policy, heatmap overlain on top-down view visualizes probability of successful pickup as a function of initial location.
(C) Representative filmstrip of behavior in the warehouse task from egocentric and side view.}
\label{fig:warehouse_performance}
\end{figure}

\begin{figure}[t]
\centering
\includegraphics[width=1.\linewidth]{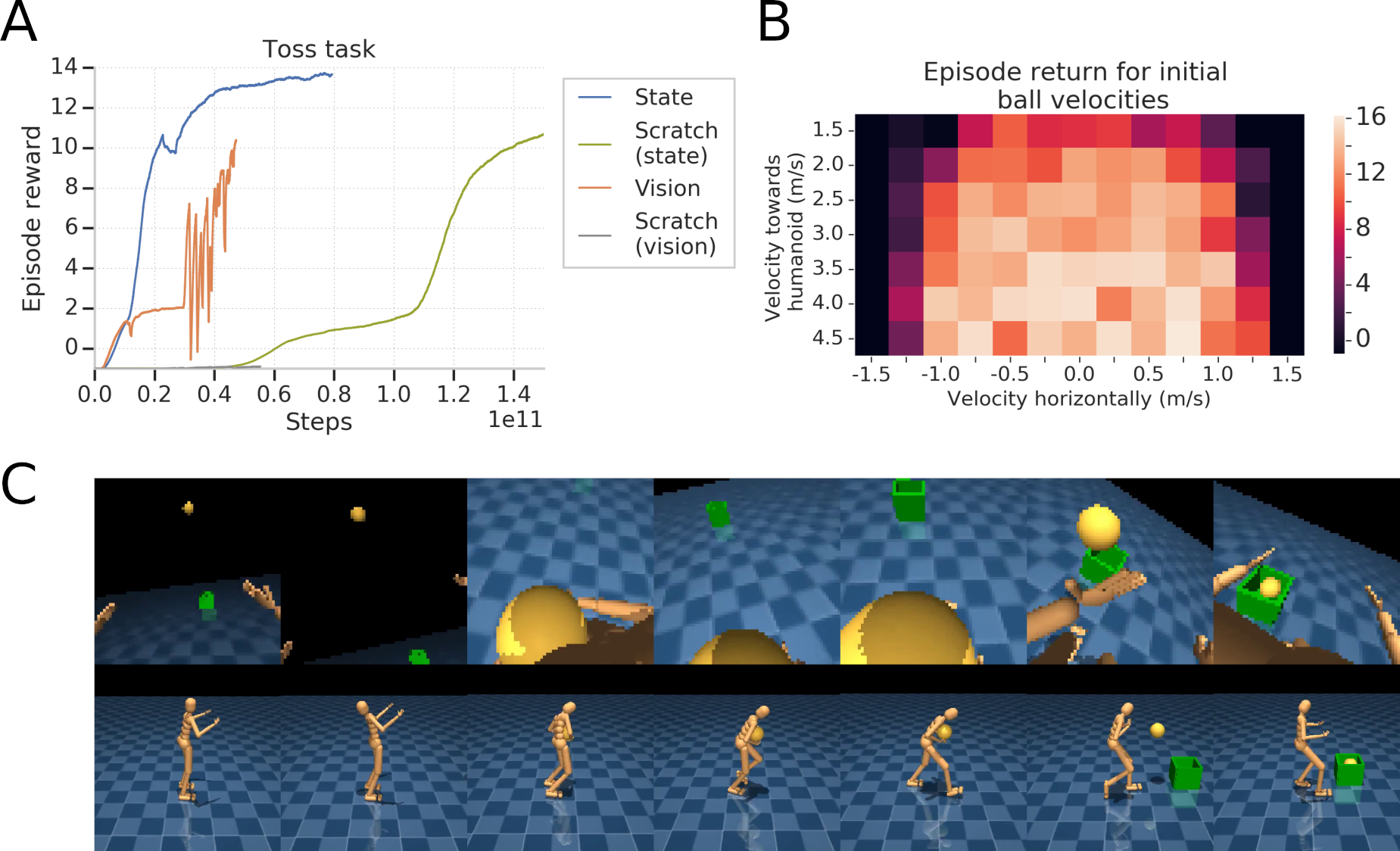}
\caption{{\bf Performance for the ``toss'' task:} (A) Representative learning curves (best of 3 seeds) comparing vision-based and state-based performance on the toss task, as a function of learner update steps.
(B) For the trained state-based policy, heatmap indicates the episode return as a function of initial ball velocity.
(C) Representative filmstrip of behavior in the warehouse task from egocentric and side view.}
\label{fig:toss_performance}
\end{figure}

We train task policies that operate from state and visual observations on both tasks and found that successful reuse is possible using either observation type.  However, note that comparable experiments from vision require longer walltime, since rendering image observations slows the simulation. On the warehouse task, visual information seems to improve learning (Figure \ref{fig:warehouse_performance}A), whereas state information is better on the toss task (Figure \ref{fig:toss_performance}A).  
{This discrepancy between the tasks is perhaps explicable -- box interaction from state may be performance limited by the available features.  Indeed, given that state features only consist of center-of-box position and orientation, precise movements that require sensory access to the edges and faces of the box relative to body and hands may be more natural to learn from visual inputs (from which this information may be more apparent).  However, in the toss task, the same state features may be more adequate for an optimal policy.}
Nevertheless, both policies using either feature set trained to a reasonable performance level.  For representative performance and behavior of the vision based policies, see the representative ``warehouse'' task 
{\color{blue} 
\href{https://youtu.be/lIXQpDyfuhs}
{\bf Video~4}} and ``toss'' task 
{\color{blue} 
\href{https://youtu.be/nysqcW6crrY}
{\bf Video~5}}.  When task policies operate from vision, the agent must learn to coordinate the body to interact with the objects that it senses and interprets from the image inputs.  
{In the warehouse task, the body and head movements lead to a somewhat jerky egocentric camera, but evidently the policy is still able to leverage this visual stream for successful object interaction.  In the toss task, the camera control is more stable and it seems intuitive that the visual stream can support object tracking for interaction, though perhaps with some ambiguity of ball size vs ball distance.
For all results in this work, any object state estimation or gaze control is learned implicitly, without any additional structure to encourage its emergence.}

Note that without reusable motor skills, an alternative is to learn the task from scratch.  This is difficult, as rewards in these tasks are sparse and therefore do not very strongly shape behavior.  And critically, in these tasks it would be very difficult to design dense rewards that incentivize the right kind of structured behavior. 
This being said, it did turn out to be possible to learning from scratch on the toss task from state information.  After training for an order of magnitude longer ($>$100e9 learning steps), the early terminations with penalty and shaping reward are sufficient to produce a policy that could solve the toss task, albeit without actually catching the ball, essentially using its back as a paddle.  This same behavior was consistently learned across multiple seeds, see {\color{blue} 
\href{https://youtu.be/Nm10vlIpiZo}
{\bf Video~6}}.  For the warehouse task, training from scratch for equivalently long did not yield behavior that could solve a whole cycle of the task, but some progress was made for certain initial conditions, see {\color{blue} 
\href{https://youtu.be/OscnI_z_tWE}
{\bf Video~7}}.  Note that for experiments from vision (slower than experiments from state), training a policy for 150e9 steps took roughly 3 weeks of wall-clock time, so it was not feasible to systematically explore training from scratch for significantly longer intervals.  

For the warehouse task, we provide an additional evaluative visualization of the final performance to provide a clearer sense of the quality of the learned solution.  For any stage of the behavior, we can take a trained agent and assess how reliably it can perform a given behavior from different initial positions.  We defined a $9\times9$ grid of initial x-y locations in the plane.  For each location we initialized the humanoid there 10 times, randomizing over orientation, body configuration (sampled from motion capture), and initial velocity.  We then computed the fraction of trials for which the humanoid was able to successfully pick up a prop.  We visualize a top down view, with the agent aiming to pick up the prop located on the pedestal on the right side of the top down view, with the heatmap of success probability overlain (Figure \ref{fig:warehouse_performance}B).  The agent is generally robust to initial position of the humanoid; only a limited fraction of initializations are too close to the pedestal and lead to failures, presumably due to initial poses or velocities that make it especially difficult.

For the toss task, we similarly wanted to provide a statistical description of the core behavior of the trained agent.  We discretized the space of initial ball velocities (both towards the humanoid and horizontally relative to the agent) -- consistent with training, we computed an initial vertical velocity such that ball would be approximately shoulder height when near the initial position of the humanoid.  We initialized the ball velocity for 10 repeats in each bin of the velocity, randomizing over other variations.  The heatmap depicted in Figure \ref{fig:toss_performance}B indicates the ``strike zone'' of parameters for which the agent is able to catch the ball.  Naturally, for initial velocities that are too horizontal it is simply not possible to catch the ball and probability of success falls off to zero (indicated by episode return of $-1$, corresponding to the ball hitting the ground).

We also remark that visual ``quality'' does not entirely align with performing the task optimally.  Throughout the course of our research, we noticed that slightly worse optimizers or termination partway through training resulted in policies that are more conservative, scored fewer points, but might subjectively be considered to look more humanlike (less hurried).  This is consistent with the humanlikeness of the movements being determined largely by the low-level controller, but performance of the task becoming increasingly hectic as the task policy ultimately controls the body to move faster and with extreme movements to achieve more reward.  For an example of a policy trained for less time, see {\color{blue} 
\href{https://youtu.be/gqH6MvabdtE}
{\bf Video 8}}.

\subsection{Task variations}
\label{sec:ablations}

In addition to demonstrating performance on the core tasks, a few themes emerged in developing our approach, for which we provide illustrative examples.  In particular, some trends that we observed include: (1) the ratio of expert skills in the NPMP matter, (2) the initializations at different phases of the task matter for the warehouse task, but aren't required for the toss task, \& (3) more extreme variations benefit from a curriculum via variations (a similar result is reported in \cite{heess2017emergence}).  

\begin{figure}[b!]
\centering
\includegraphics[width=1.\linewidth]{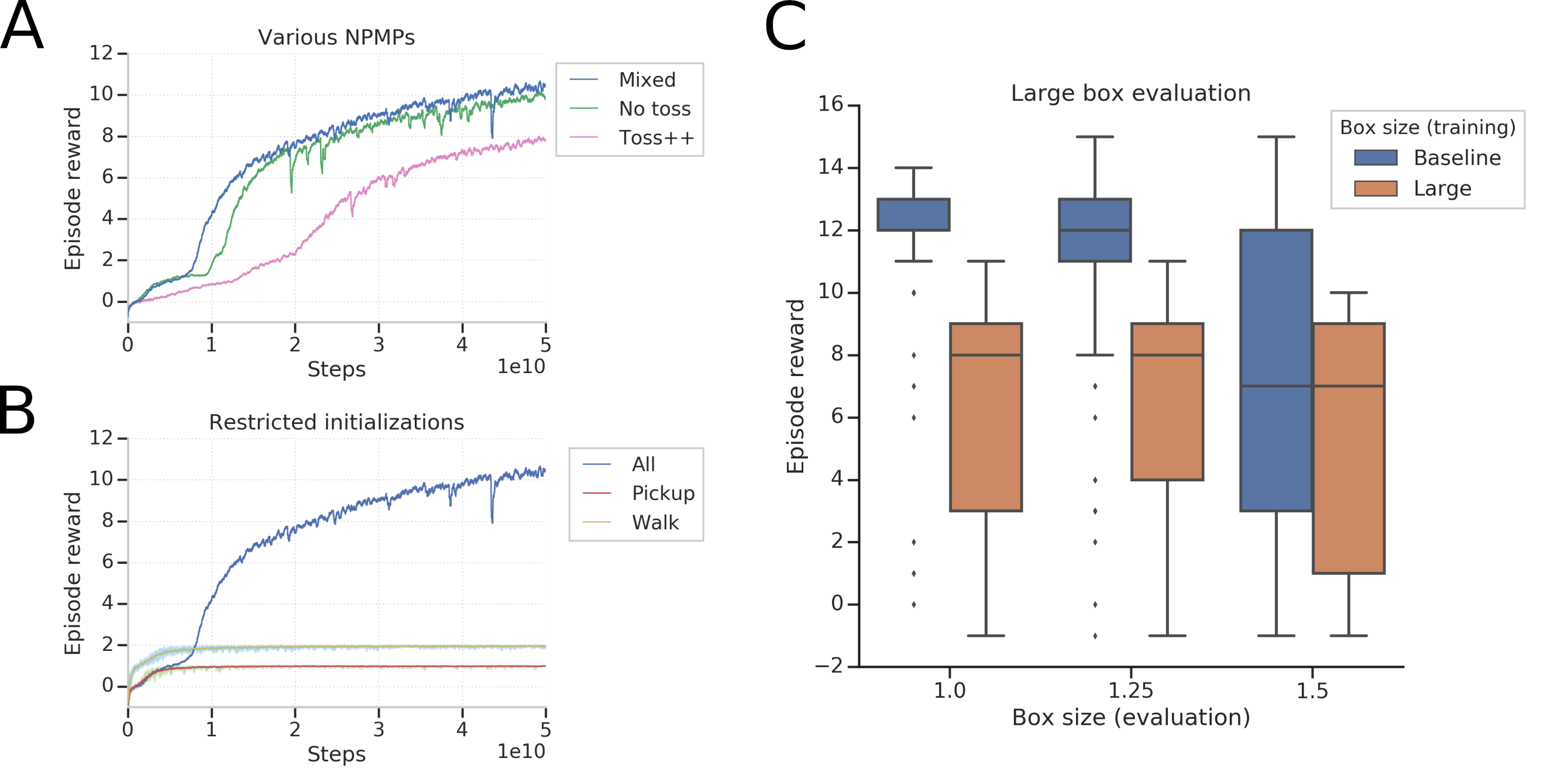}
\caption{{\bf Ablations and probes.} (A) Performance on the warehouse task as a function of using NPMPs trained with different ratios of expert data. ``Mixed'' uses natural proportions of all of our warehouse and ball tossing data, ``no toss'' omits all tossing data, and ``toss++'' uses ball toss experts upsampled by a factor of two. 
(B) Comparison of performance on the warehouse task under the default setting involving initializations at ``all'' phases of the task versus training with initializations restricted to either the ``pickup'' or ``walk'' phases.
(C) We trained the task policy to perform the warehouse task with either the baseline variation in box sizes (blue) or only a smaller range of large boxes (orange) and then evaluated performance only interacting with large boxes.  Variation on the wider distribution during training improved performance on the evaluation tasks. }
\label{fig:warehouse_variations}
\end{figure}

First, we consider how important the relative ratios of different skills are in the NPMP.  In extreme cases, this is trivially important.  For example, an NPMP module that only contained locomotion skills, without object interactions, would intuitively offer limited utility for transfer to the warehouse task.  A more nuanced question is how important the relative quantities of ball tossing behavior versus warehouse behavior affect the ability of the NPMP to learn the two tasks.  For illustration, we trained three NPMPs, one that only had access to warehouse experts, one that had both warehouse experts and ball toss experts in proportion to how much was collected (more motion capture was warehouse relative to toss demonstrations), and one that trained on twice as much data from toss experts -- that is, when training the NPMP, we recorded twice as many trajectories from ball toss experts as we did for other experts, thereby over-representing these experts in the NPMP training data.  Note that in the toss upsampled NPMP, toss experts are over-represented relative to our motion capture, but there was still more warehouse data relative to toss data even with this upsampling. We observed that while the upsampled toss NPMP learned an arguably slightly more aesthetically satisfying toss behavior (no meaningful change in performance), it was more difficult for the upsampled toss NPMP to learn the warehouse task.  In figure \ref{fig:warehouse_variations}A, we show comparisons of these different NPMPs on the warehouse task.  While ultimately, the upsampled toss NPMP was able to learn the warehouse task, it was consistently lower and less robust for other hyperparameters.

In addition to the balance of expert data, we also examined the need to initialize the behavior in different phases of the warehouse task.  In the warehouse task, the training episodes are initialized in all phases of the task in poses sampled from motion capture, forming a curriculum over variations.  In the toss task, we did not initialize episodes in different task phases.  To illustrate the need for starting the warehouse task in the various phases, we ran comparisons involving starting only in the pickup or walk phases of the task and found that neither of these are able to learn to solve the task (see figure \ref{fig:warehouse_variations}B).  

We also explored both decreasing and increasing the range of procedural variations across episodes.  Based on previous work \citep{heess2017emergence}, it was our starting intuition to design the task with a sensible range of variations to facilitate learning -- this meant that our initial distribution of variations basically worked.  However, we also attempted to train the task policy to perform the task with only large boxes.  We probed the original task policy trained on variable box size on variants of the warehouse task that only included larger boxes, and we see that training with variations improves performance on the probe task (figure \ref{fig:warehouse_variations}C).  Essentially, it was much more difficult to learn to solve this task without the variations in box size -- no policy fully solved the task.  For representative failure mode behavior, see {\color{blue}
\href{https://youtu.be/SmJ6DshlpJQ}
{\bf Video 9}}. 

Finally, we considered training on a wider distribution than our standard range of pedestal heights and this tended to work -- this indicates that a broader, continuous task distribution could allow a policy to perform a wide range of movements, so long as exploration and learning are guided from some examples that are initially achievable.  See a representative {\color{blue}
\href{https://youtu.be/M-ddvzRjWXc}
{\bf Video 10}} showing performance when trained on this broader range of pedestal heights, including pedestals that are quite low to the ground as well as higher up.

\section{Discussion}

In this work, we demonstrated an approach for transfer of motor skills involving whole body humanoid movement and object interaction.  We showed that a relatively small set of demonstration data can be used to provide a fairly generic low-level motor skill space that can be leveraged to improve exploration and learning on various tasks that can be solved via movements similar to those in the expert demonstrations.  Importantly, a single skill module is multipotent, permitting reuse on multiple transfer tasks.  The resulting task controllers can follow high-level goals and adapt naturally to changes or interactive perturbations of the scene and object details (see {\color{blue} 
\href{https://youtu.be/2rQAW-8gQQk}
{\bf Video 1}}).  
{These interactions with trained controllers are enabled by the ability to deploy trained controllers in real-time on a standard computer.}
Critically, a physics-based controller always responds to a scene in a manner that is consistent with the physics of the world, unlike kinematic approaches which can generalize in physically impossible ways. 
Furthermore, the use of egocentric vision induces natural behavioral phenomena such as directing the gaze towards objects of relevance; but perhaps more significantly, it makes task-specific scene representations unnecessary. The agent can understand what to do by looking at the scene.

What differentiates our approach from most preceding work that leverages demonstrations for physics-based control is that we synthesize a single, multipotent skill module.  Instead of having to stay close to demonstrations of a single object interaction, we provide a large set of unlabeled demonstrations and automatically generalize skills from them.  One open question that we believe will be important in future efforts involves how to best trade off the specificity of exploration, provided by staying close to a narrow set of demonstrations, versus generality obtained through leveraging more diverse demonstrations.  
As we showed, there is still some sensitivity to the relative ratios of the various skills in the space. It may be fundamental that there is some trade-off between exploration guidance and generality, and it would be interesting to better understand this in high-dimensional control settings.  

A similar trade-off currently exists between generality and visual quality. We note that in this work, the task objectives are quite basic -- essentially providing sparse indications of task progress. Where the controller generalizes beyond the motion capture, the behavior may look slightly less natural.  We are optimistic that further refinement of the movement appearance could result from additional objectives {(e.g. smoothness or energy regularization)} when training the task policies, additional data that helps more completely cover the space of relevant behaviors, and adjustments to the physics of the body.

We have outlined a generic architectural scheme for motor reuse, within which it is possible to implement the encoder and decoder using various specific neural networks. We have elected to use the simple choice of shallow MLPs (both our encoder and decoder have two hidden layers). 
Conceptually, we view the specific choice of network as an important, but secondary, consideration in the context of the overall approach. Network engineering may further improve this scheme, for instance through either faster learning or in terms of motion quality.
In particular, \citealt{peng2019mcp} have proposed an alternative architecture that could be applied in this setting involving additional multiplicative structure and composable primitives. Although \citealt{peng2019mcp} reported that their multiplicative compositional policies (MCP) network structure performed better than an MLP in their settings, our own exploratory investigation of the reported MCP architecture (using 8 primitives) did not reproduce this benefit.  Instead, we found that while MCP reuse did produce generally similar movements in early stages of learning, it neither took off faster nor achieved effective performance in the warehouse setting.  We speculate that this difference may have to do with the necessary degrees of freedom for the task -- in the warehouse setting, limiting the reusable control to compositions of a discrete set of primitives may impair their reusability outside of the movements that are highly represented in the reference data.  Regardless, as a comparison of decoder architectures is not a focus of this work, we did not pursue this deeply, and we leave more systematic comparison of architectural choices to future work.

There are a few additional caveats concerning the present approach that are related to the difficulty of exploration and learning in complicated tasks.
We train the task policies by model-free RL, and this is not efficient with respect to data (requiring many environment interactions and learner updates).  Substantially, this slowness arises from dithering exploration at the level of the task policy.
While we use low-level skills to structure exploration, this alone is not sufficient to learn to control high-dimensional bodies interacting with objects, especially from only sparse rewards. We are optimistic that future improvements will accelerate learning, perhaps partly through more intelligent, goal-directed exploration strategies. For example, the motif of using skills for exploration could potentially be repeated hierarchically to encourage richer behavior.
Another limitation is that, for the warehouse task, we leverage a curriculum via informative motion capture initializations, which expose the agent to favorable states that it may not have discovered on its own.
It is interesting to note that the use of initializations is not required for the ball toss, where the combination of the ball being thrown towards the humanoid (forcing it to engage) and a weak shaping reward to induce movement towards the box are adequate.

Taken together, these limitations restrict the present approach to simulation settings; however there is a growing literature on approaches involving transfer of policies trained in simulation to real world systems (sim-to-real) \citep{rusu2017sim, sadeghi2016cad2rl, tobin2017domain, andrychowicz2018learning, zhu2018reinforcement, tan2018sim, hwangbo2019learning, xie2019iterative}.  
While we believe that the study of sophistocated motor control problems in simulation is an important area of research in its own right, sim-to-real may offer a path to translate these results into real world applications.

\begin{acks}
We thank Tim Lillicrap for constructive input at the outset of the project, Vicky Langston for help coordinating the motion capture acquisition, Thomas Roth{\"o}rl for assistance during our studio visit, and Audiomotion Studios for services related to motion capture collection and clean up.  We also thank others at DeepMind for input and support throughout the project.  Finally, we thank Jaakko Lehtinen for comments on the manuscript.
\end{acks}

\bibliographystyle{ACM-Reference-Format}
\bibliography{references}

\appendix
\section*{Appendices}
\renewcommand\thefigure{A.\arabic{figure}}    
\setcounter{figure}{0}

\section{Simultaneous tracking and calibration}
\label{stac_details}
Simultaneous tracking and calibration (STAC) is an algorithm for inferring joint angles of a body from point-cloud data when it is not known in advance precisely where the markers are on the body \citep{wu2013stac}.  The relevant variables include the body which has a pose ($\vec{q}$) as well as marker positions that are fixed to it ($\vec{x}_m$).  We observe via motion capture the sensor readings ($\vec{s}^{\,\star}$) which should be equal to the positions of the markers at each timestep, up to negligible noise. STAC makes the assumption that the markers are rigidly attached to the body with fixed offsets ($\vec{x}_m$) -- if those offsets are known, a forward kinematics call ($f_k(\cdot)$) allows us to compute the positions at which we expect sensor readings.

For known marker offsets, the pose of the body ($\vec{q}$) can be inferred by optimizing (per frame):
\begin{equation}
    \argmin_{\vec{q}} ||f_k(\vec{q}, \vec{x}_m) - \vec{s}^{\,\star}||^2_2
\end{equation}

We additionally know that the marker offsets should be the same at every timestep (assuming rigid attachment of markers).  So similarly, if the pose of the body is known, the marker offsets can be inferred by optimizing:
\begin{equation}
    \argmin_{ \vec{x}_m} \sum_i ||f_k(\vec{q}_i, \vec{x}_m) - \vec{s}^{\,\star}_i||^2_2
\end{equation}

So overall, to perform joint optimization over unknown marker offsets and poses, we alternate between these optimization steps.  We initialize the pose of the body to the null pose (approximately a t-pose) and roughly initialize the marker offsets by placing markers on the body part, without precise tuning.  The first optimization is of the pose, using the initial, coarsely placed markers (per frame).  We then optimize the marker positions using frames sampled at a regular interval throughout a range-of-motion
{\color{blue} 
\href{https://youtu.be/WowU5w12Aa8}
{\bf Video 11}}.  We then re-optimize the joint angles per frame.  We found that further alternation was not required and the marker offsets that are found using the range-of-motion clip worked well for all other clips.

In practice we also use a small regularization term, encouraging joints angles to be near the null pose, and we also warm start the per-frame optimization at the inferred pose from the preceding timestep.

\section{Single-clip tracking objective}
\label{tracking_details}
In the main text, we described that the tracking reward arises from a weighted sum of terms that score how well different features of the reference are being tracked.  More specifically, these objectives are:
\begin{align*}
E_{\rm ori}&= ||\log(\vec{q}_{\rm ori} \cdot \vec{q}^{\, \star -1}_{\rm ori})||_2 \\
E_{\rm gyro}&= 0.1\cdot||\vec{q}_{\rm gyro} - \vec{q}^{\, \star}_{\rm gyro} ||_2 \\
E_{\rm obj} &= ||\vec{x}_{\rm obj} - \vec{x}^{\, \star}_{\rm obj}||_2
\end{align*}
\begin{align*}
E_{\rm qpos} &= \frac{1}{N_{\rm qpos}}\sum|\vec{q}_{pos} - \vec{q}^{\, \star}_{pos}| \\
E_{\rm qvel}&= \frac{1}{N_{\rm qvel}}\sum|\vec{q}_{\rm vel} - \vec{q}^{\, \star}_{\rm vel}| \\
  E_{\rm app} &= \frac{1}{N_{\rm app}}\sum||\vec{x}_{\rm app} - \vec{x}^{\, \star}_{\rm app}||_2 \\
E_{\rm vel}&= 0.1\cdot \frac{1}{N_{\rm vel}}\sum|\vec{x}_{\rm vel} - \vec{x}^{\, \star}_{\rm vel}| 
\end{align*}

where $\vec{q}$ represents the pose or velocity and $\vec{q}^{\, \star}$ represents the reference value.  The $\vec{x}_{\rm app}$ and $\vec{x}_{\rm obj}$ are 3D Cartesian vectors from the root to the various appendages (head, hands, feet) or object (box or ball) in the root frame.  The root is located in the pelvis of the humanoid. $\vec{x}_{\rm vel}$ is in the global reference frame.  
In this work, for the body terms, we used coefficients 
$w_{\rm qpos}=5$,
$w_{\rm qvel}=1$,
$w_{\rm ori}=20$,
$w_{\rm app}=2$,
$w_{\rm vel}=1$,
$w_{\rm gyro}=1$. This has been used in previous work \citep{merel2018hierarchical}.  The object tracking term coefficient, new to this work, was tuned to $w_{\rm obj}=10$ to relative strongly enforce object tracking.  The same values are used for all clips, despite the diversity of behaviors, indicated relative robustness of this approach.

\section{One-shot imitation evaluation}
\label{one_shot}

\begin{figure}[h!]
\centering
\includegraphics[width=1.\linewidth]{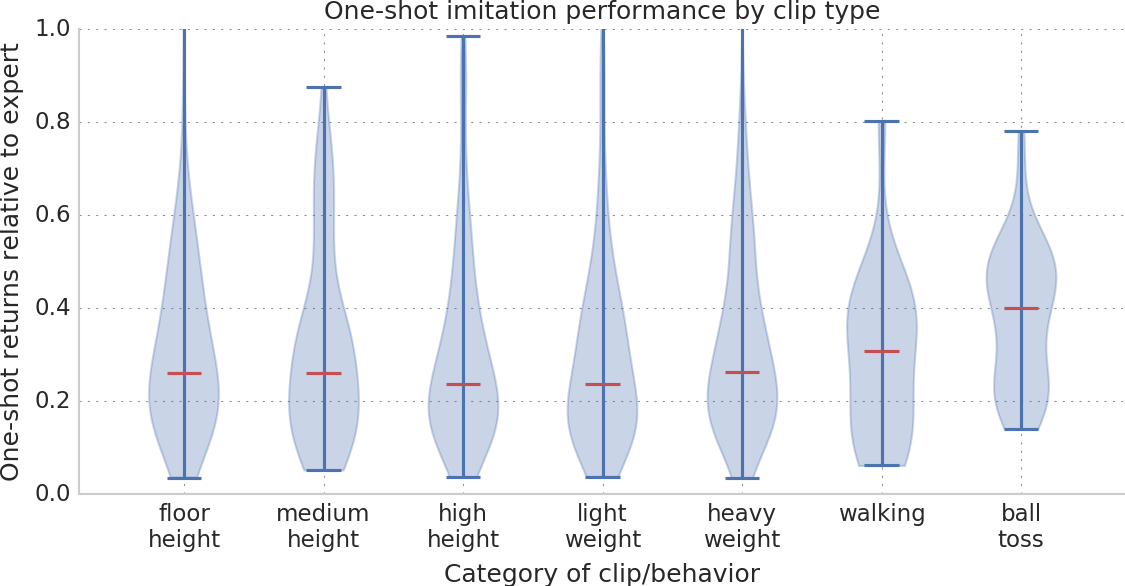}
\caption{{\bf One-shot analyses:} Here we depict, by behavior category, the one-shot performance of the trained NPMP.  Note that relative performance $>1$ happens if an expert is imperfect and the one-shot imitation is effectively denoised. 
}
\label{fig:one_shot}
\end{figure}

One-shot imitation involves providing the trained NPMP with a state-sequence and asking it to generate a sequence of actions that would reproduce that movement.  In asking the trained NPMP to perform one-shot imitation, we get a glimpse into which skills it is able to perform well, and we can be assess this performance for overlapping subcategories of clips.  Note that one-shot imitation is not actually the objective that the NPMP was trained to perform, and one-shot imitation is difficult due to object interactions (see figure \ref{fig:one_shot}).
Both walking behavior and ball toss behavior are better captured than the pickup and putdown interactions with boxes.  This presumably reflects the fact that in terms of timesteps of data, there are fewer moments at which the difficult box interactions are being performed.  As such, these quantifications may leave a misleading impression that one-shot behavior is worse than it is.  To complement these quantitative metrics, we also provide a {\color{blue} 
\href{https://youtu.be/KDiiCS9H2-Q}
{\bf Video 12}} showing a representative assortment of one-shot behavior which show that while the object interactions can be difficult, movements are broadly sensible.

\section{Instructed warehouse task details}
\label{warehouse_supp}
The warehouse task rewards moving a box from one pedestal to another, and repeating this process. The environment consists of a flat ground with four pedestals and two boxes that can be moved freely (note that we varied the number of boxes and pedestals as well, not reported in this paper, and results are similar). The distance of each pedestal from the origin is individually drawn from a uniform distribution between 2.5 and 3.5 meters, and the pedestals are at equispaced angles around the origin. The height of each pedestal is set randomly from between 0.45 and 0.75 meters.
The size of each box is taken from one of our motion capture trajectories, but with a random multiplicative variation of between 0.75 and 1.25 applied. The mass of each box is also individually drawn from a uniform distribution between 2kg and 7kg (the real boxes are either 3kg or 10kg). The size and mass of each box is not directly provided as an observation to the agent.

This task can be logically divided into four phases: walk empty-handed to a pedestal (GOTO), lifting the box from a pedestal (LIFT), carrying the box to a different pedestal (CARRY), and putting it down on the target pedestal (PUTDOWN). In our current work, we provide the agent with an observation that tells it which of these four phases it should be pursuing at a given timestep, as a one-hot vector. The position of the \textit{focal pedestal} and \textit{focal box} relative to the walker is also provided as observations, where the \textit{focal box} is the box that needs to be moved, and the \textit{focal pedestal} is dependent on the phase of the task: in GOTO and LIFT it is the pedestal on which the box is initially placed, while in CARRY and PUTDOWN it is the target pedestal. 
Each of the four phases has well-defined success criteria, as detailed in Table \ref{table:phases} (an empty cell indicates that a particular type of criterion is not used to determine success of a phase).

\begin{table}[htb]
\centering
\begin{tabular}{|p{1.6cm}|p{1.8cm}|p{1.8cm}|p{1.8cm}|}\hline
    \textbf{Phase} & \textbf{Walker position} & \textbf{Walker/box contact} & \textbf{Pedestal/box contact} \\\hline
     GOTO          & within 0.65 meter of focal pedestal & & \\\hline
     LIFT          & & at least one contact point with each hand & no contact points \\\hline
     CARRY         & within 0.65 meter of focal pedestal & at least one contact point with each hand & \\\hline
     PUTDOWN       & & no contact points & at least 4 contact points \\\hline
\end{tabular}
\caption{The four phases of the warehouse task.}
\label{table:phases}
\end{table}

At each timestep, task logic determines whether the agent has successfully completed its current phase. If it has, a reward of 1.0 is given at that timestep only, and the task is advanced to the next phase. The phase transition is determined by a simple state machine:
$$\mathrm{GOTO} \rightarrow \mathrm{LIFT} \rightarrow \mathrm{CARRY} \rightarrow \mathrm{PUTDOWN} \rightarrow \mathrm{GOTO} \rightarrow ...,$$
and the task repeats indefinitely up to a final episode duration (15s simulated time), at which point the episode is terminated with bootstrapping.  While there is no prespecified maximum score, obtaining an undiscounted return greater than 10 within a 15s episode requires moving through the phases rapidly.  Note that the episode is terminated with a failure (no bootstrapping) if either the walker falls (contact between a non-foot geom of the walker and the ground plane) or if a box is dropped (contact between one of the boxes and the ground plane).

At the beginning of each episode, after randomly sampling the variations described above, one of the four phases is sampled uniformly as the initial phase for the episode. A motion capture trajectory is picked at random, and a random timestep from the clip consistent with the active phase in the episode is sampled. 
{Note that the motion capture clips can be automatically segmented into phases by simply applying the task logic to the motion capture reference, so no manual segmentation is required (beyond defining the task).}
The joint configuration of the walker, the position of the box relative to the walker, and both the walker's and box's velocities, are synchronized to the state from this motion capture timestep. If the episode begins in either the LIFT or PUTDOWN phase, the displacement of the walker from the focal pedestal is also synchronized, otherwise we apply a random translation and rotation around the $z$-axis (i.e. yaw) to the walker and prop together as a rigid body.

\section{Ball toss task details}
\label{toss_supp}

{The toss task encourages catching a ball and subsequently throwing it into a bucket.  The initial pose of the walker is randomly sampled from a range of motion capture poses related to ball tossing.  The ball size and mass are procedurally randomized (radius randomly multiplied by a factor uniformaly sampled between .95 and 1.5; mass sampled uniformly from the range of 2 to 4kg) and the angle and velocity of the ball are also procedurally randomized such that the ball is generally ``thrown'' towards the humanoid.  More precisely, the ball is always initialized in middair, behind the bucket, at a distance of roughly 3m from the humanoid ($d_{x}$).  To initialize the ball velocity in a way that ensures it is projected towards the humanoid in an appropriate \textit{strike zone}, we must determine initial 3D velocity components of the ball, which define its trajectory. We first pick a random velocity towards the humanoid ($v_{x}$ between 1.5 and 4.5m/s).  We can also select a random horizontal velocity relative to the walker ($v_y$ between .75m/s leftwards or rightwards).  For the vertical component, we compute a random target height ($d_z$ between .1 and .4m from the ground), and then we analytically compute the time at which the ball should hit the humanoid ($t_{hit}=d_{x}/v_{x}$) as well as the initial vertical velocity required to hit the randomly selected target ($v_z = (4.9t_{hit}^2 + d_z)/t_{hit}$). For robustness, random angular velocities are also applied to the ball at the initial timestep.} 
    
The incentives of the task are specified through rewards and termination logic. The primary element of the task is that if the ball touches the ground or if the humanoid falls (contact between a non-foot geom of the walker and the ground plane), the episode terminates with a negative reward.  This strongly disincentivizes letting the ball fall to the ground and encourages the humanoid to remain standing.  Even reliably achieving this level of performance over the range of procedural ball trajectories is difficult.  In addition, once the ball reaches the humanoid, a shaping reward is activated that corresponds to a small positive per-timestep reward inversely related to the distance between the ball and the bucket (in the x-y plane, neglecting vertical height).  This reward encourages the humanoid, after catching the ball to walk towards the bucket.  Finally, if the ball is in the bucket, there is a moderate per-timestep reward encouraging dropping the ball into the bucket -- this final reward is sparse in the sense that it is achieved iff there is a contact between the bottom of the bucket and the ball.  Once the agent has learned to drop the ball into the bucket, it learns to do this earlier (i.e. throw the ball) to achieve the reward as soon as possible.

\section{Supplementary video captions}
\label{videos}

\textbf{Video 1} Overview video summarizing highlights of the paper.

\noindent\textbf{Video 2}: Kinematic playback of a motion capture clip of a box interaction.

\noindent\textbf{Video 3}: Kinematic playback of a motion capture clip of ball tossing.

\noindent\textbf{Video 4}: A representative illustration of the behavior of a successfully trained vision-based policy on the ``warehouse'' task.

\noindent\textbf{Video 5}:  A representative illustration of the behavior of a successfully trained vision-based policy on the ``toss'' task.

\noindent\textbf{Video 6}: A representative illustration of the behavior learned by a policy trained from scratch on the ``toss'' task (from state).

\noindent\textbf{Video 7}: A representative illustration of the behavior learned by a policy trained from scratch on the ``warehouse'' task (from vision).

\noindent\textbf{Video 8}: A representative illustration of the behavior of a partially trained vision-based policy on the ``warehouse'' task.

\noindent\textbf{Video 9}: A representative illustration of the behavior learned on the ``warehouse'' task when training only with large boxes.

\noindent\textbf{Video 10}: A representative illustration of the behavior of a successfully trained vision-based policy on the ``warehouse'' task when trained on a wider range of pedestal heights.

\noindent\textbf{Video 11}: Kinematic playback of a range-of-motion motion capture clip used for STAC calibration.

\noindent\textbf{Video 12}: Examples of one-shot imitation of object interaction behaviors.

\end{document}